\documentclass[10pt,twocolumn,letterpaper]{article}

\usepackage{wacv}              % To produce the CAMERA-READY version
\usepackage{graphicx}
\usepackage{amsmath}
\usepackage{amssymb}
\usepackage{booktabs}
\usepackage{array}

\usepackage{xpatch}

\makeatletter
\newinsert\copyrightnote@ins
\count\copyrightnote@ins=1000
\dimen\copyrightnote@ins=8in
\newcommand*\copyrightnote@hook
  {%
    \unvbox\copyrightnote@ins
    \global\let\@makecol\copyrightnote@makecol
  }
\let\copyrightnote@AtBeginDocument\AtBeginDocument
\AtBeginDocument{\let\copyrightnote@AtBeginDocument\@firstofone}
\newcommand*\copyrightnote@firstuse
  {%
    \gdef\copyrightnote@firstuse
      {\global\skip\copyrightnote@ins=\bigskipamount\gdef\copyrightnote@firstuse{}}%
    \global\let\copyrightnote@makecol\@makecol
    \xpatchcmd\@makecol{\unvbox\footins}{\unvbox\footins\copyrightnote@hook}
      {}{\GenericError{}{patching @makecol failed}{}{}}
    \copyrightnote@AtBeginDocument
      {%
        \ifvoid\footins
          \insert\footins{}% 
        \else
          \global\skip\copyrightnote@ins=\bigskipamount
        \fi
      }%
  }
\newcommand\copyrightnote[1]
  {%
    \copyrightnote@firstuse
    \copyrightnote@AtBeginDocument
      {%
        \insert\copyrightnote@ins
          {%
            \reset@font\footnotesize
            \interlinepenalty\interfootnotelinepenalty
            \splittopskip\footnotesep
            \splitmaxdepth\dp\strutbox
            \floatingpenalty\@MM
            \hsize\columnwidth
            \@parboxrestore
            \color@begingroup
            \parindent1em
            \noindent
            \hskip1.8em
            \ignorespaces #1\@finalstrut\strutbox
            \color@endgroup
          }%
      }%
  }
\makeatother

\usepackage[accsupp]{axessibility}  % Improves PDF readability for those with disabilities.

\usepackage[pagebackref,breaklinks,colorlinks]{hyperref}

\usepackage[capitalize]{cleveref}
\crefname{section}{Sec.}{Secs.}
\Crefname{section}{Section}{Sections}
\Crefname{table}{Table}{Tables}
\crefname{table}{Tab.}{Tabs.}

\def\wacvPaperID{2252} % *** Enter the WACV Paper ID here
\def\confName{WACV}
\def\confYear{2024}

\begin{document}

\title{Joint Depth Prediction and Semantic Segmentation with Multi-View SAM}

\author{Mykhailo Shvets\\
University of North Carolina\\
Chapel Hill\\
{\tt\small mshvets@cs.unc.edu}
\and
Dongxu Zhao\\
University of North Carolina\\
Chapel Hill\\
{\tt\small dongxuz1@cs.unc.edu}
\and
Marc Niethammer \\
University of North Carolina\\
Chapel Hill\\
{\tt\small mn@cs.unc.edu}
\and
Roni Sengupta\\
University of North Carolina\\
Chapel Hill\\
{\tt\small  ronisen@cs.unc.edu}
\and
Alexander C. Berg\\
University of California\\
Irvine\\
{\tt\small bergac@uci.edu}
}
\maketitle
\copyrightnote{To appear in the 2024 IEEE/CVF Winter Conference on Applications of Computer Vision \copyright~2023 IEEE \\
Personal use of this material is permitted. Permission from IEEE must be obtained for all other uses, in
any current or future media, including reprinting/republishing this material for advertising or promotional
purposes, creating new collective works, for resale or redistribution to servers or lists, or reuse of any
copyrighted component of this work in other works.
}

%%%%%%%%% ABSTRACT
\begin{abstract}

    Multi-task approaches to joint depth and segmentation prediction are well-studied for monocular images. Yet, predictions from a single-view are inherently limited, while multiple views are available in many robotics applications. On the other end of the spectrum, video-based and full 3D methods require numerous frames to perform reconstruction and segmentation. With this work we propose a Multi-View Stereo (MVS) technique for depth prediction that benefits from rich semantic features of the Segment Anything Model (SAM). This enhanced depth prediction, in turn, serves as a prompt to our Transformer-based semantic segmentation decoder. We report the mutual benefit that both tasks enjoy in our quantitative and qualitative studies on the ScanNet dataset. Our approach consistently outperforms single-task MVS and segmentation models, along with multi-task monocular methods.  
    
   \vspace{-1em}
\end{abstract}

%%%%%%%%% BODY TEXT

\section{Introduction}
\label{sec:intro}

Depth prediction and semantic segmentation are core tasks for visual understanding in robotic perception. The ability to recognize \emph{what} the objects are and \emph{where} they are in the scene, with respect to the robot's viewpoint, plays a key role in enabling effective navigation and interaction in complex environments.

\begin{figure}[t]
  \centering
   \includegraphics[width=0.9\linewidth]{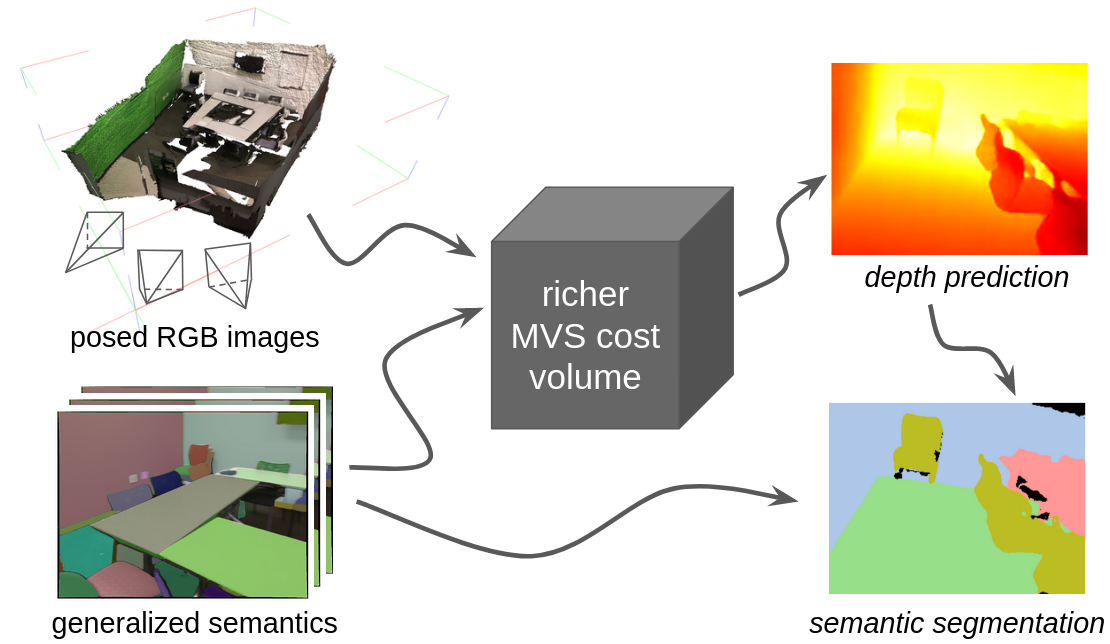}

   \caption{Generalized semantic features help to build a richer cost volume for MVS. In turn, the depth predicted from the cost volume serves as a rich prompt for semantic decoding.}
   \label{fig:conceptual}
     \vspace{-1.5em}

\end{figure}

The tremendous success of deep learning methods has significantly improved the performance of both tasks in recent years. Moreover, depth prediction and semantic segmentation have been shown to carry mutually beneficial information: depth complements the visual RGB information with geometric cues to help segmentation~\cite{hazirbas2017fusenet,jiang2018rednet,valada2020adapnetplusplus,fischedick2023efficient,zhang2022cmx}, and conversely, segmentation can help guide depth prediction~\cite{zhu2020edgeofdepth,chen2019towards,klingner2020self,guizilini2020semanticallyguided}. To take it one step further, rather than treating tasks in isolation or solving them sequentially, a range of multi-task approaches were proposed to jointly solve depth estimation and semantic segmentation, especially in the monocular image domain~\cite{nekrasov2019real,liu2019mtan,vandenhende2020mti,xu2018padnet}. The discovery of this complementarity between the tasks improves the generalization capabilities of both~\cite{vandenhende2021multi}. In the single image domain, it is natural, from an architectural design perspective, to combine these two since both can be addressed using a 2D convolutional encoder-decoder architecture. Nevertheless, a significant amount of geometric and relational context is often missing from the monocular estimation algorithms since it is challenging to estimate geometry and semantics purely from a single view.

On the other end of the spectrum, full 3D recognition~\cite{hu2021bpnet} and reconstruction~\cite{murez2020atlas,sun2021neuralrecon} models  showed impressive performance on a range of tasks. These approaches typically operate on video sequences~\cite{duzceker2021deepvideomvs} or complete 3D scans~\cite{dai20183dmv}. Such data can be expensive to acquire and/or process. Moreover, in robotics applications, scenarios arise that necessitate achieving visual understanding from just a few frames without a holistic scene analysis. This is particularly prominent in dynamic scenes that are impossible to measure completely at a single point in time.

Multi-View Stereo (MVS) approaches have the potential to bridge the gap between weaker monocular methods suffering from the absence of explicit geometry notions, and full video-based or 3D-based solutions. Indeed, the goal of MVS models is to successfully predict depth from just a few input views by leveraging principles and concepts from traditional camera geometry~\cite{hirschmuller2007stereo}. Modern MVS models~\cite{yao2018mvsnet,gu2020casmvsnet,ding2022transmvsnet} extract deep 2D geometry-aware features for cost volume construction. 
Unfortunately, while these MVS models do perform complex intra-frame reasoning, their feature extraction process is relatively simplistic. Due to this, the model predictions can suffer and may be highly error-prone, most notably in textureless regions. This challenge can only be mitigated to a limited extent with standard multi-scale and planar prior methods~\cite{xu2019multi,xu2022multi}.

We propose the use of semantic cues to help recover better from those errors. There are several challenges to overcome to achieve joint MVS depth prediction and segmentation in indoor settings.
Our first challenge is that 2D features in the MVS pipeline are not naturally good at extracting broad semantics.
A vast amount of models trained end-to-end for 2D semantic segmentation exist~\cite{deeplabv1, deeplabv2, deeplabv3, valada2020adapnetplusplus, pspnet, segnet, ronneberger2015unet} that could be used to augment MVS 2D features with semantic cues. However, in our studies we show that their intermediate features might still lack the generalizability required to help MVS. One of the main reasons is that the sample diversity in existing indoor datasets~\cite{straub2019replica,dai2017scannet} is fairly limited---even with millions of images, those datasets typically only contain a limited number of scenes.
Second, contrary to the single-task domain, performing joint prediction of MVS depth and semantics in a multi-task way is less straightforward due to the divergence of their models' architectural design: while semantics still follow the convolutional or attention-based design, MVS depth is naturally inferred from correlation or variance-based cost volumes~\cite{yao2018mvsnet}. This is the gap we plan on bridging in this work. 

\begin{figure}
  \centering
   \includegraphics[width=1.0\linewidth]{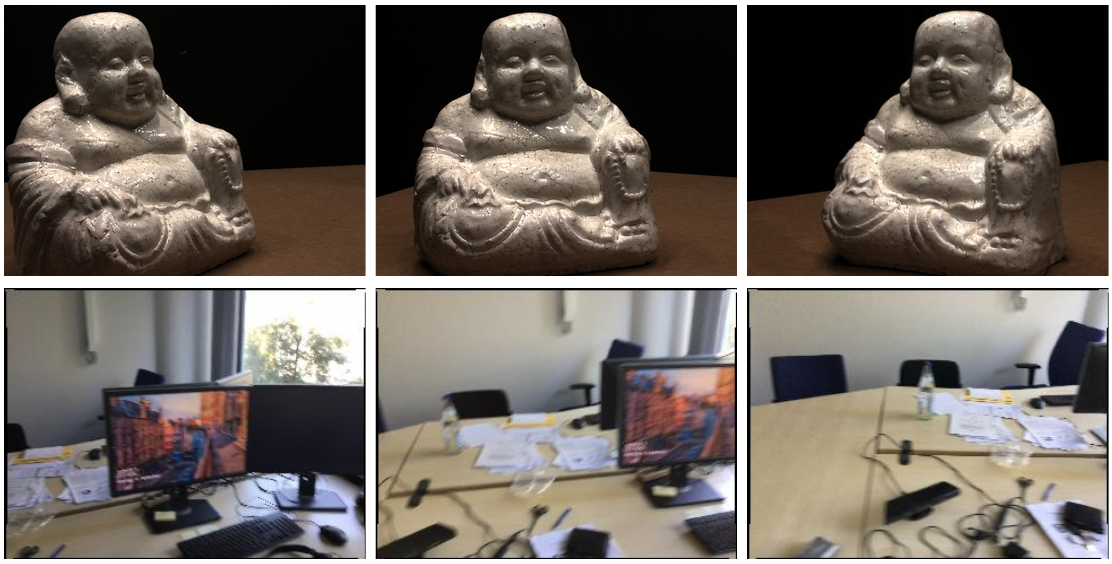}

   \caption{Challenges of view sampling in real-world scenarios. Top: DTU~\cite{jensen2014dtu} dataset, commonly used for benchmarking MVS methods, has \emph{inward} facing cameras: the views have wide baseline and the object of interest is featured from multiple angles. Bottom: ScanNet~\cite{dai2017scannet} indoor recognition dataset often features~\emph{outward} facing cameras with camera motion close to pure rotation.}
   \label{fig:dtu_scannet_views}
   \vspace{-1em}
\end{figure}

Additionally, MVS methods are not yet well explored for indoor environments and robotics scenarios. Popular datasets that are used to benchmark MVS methods~\cite{jensen2014dtu,knapitsch2017tanks} often feature inward-facing views with a single object of interest captured from many perspectives, while in the indoor scenario, it is often the case that cameras are facing outwards and thus view baseline sampling options may be limited, as shown in Figure~\ref{fig:dtu_scannet_views}. 

Our key contribution is to introduce a joint depth and segmentation prediction network using the multi-view stereo framework where we can reason about the 3D scene using a few images (3-5). The previous joint depth and segmentation approaches are either monocular~\cite{xu2018padnet,nekrasov2019real,vandenhende2020mti,liu2019mtan}, or require a full dense scan of the environment~\cite{dai20183dmv,hane2013joint}. Our pipeline's concept is illustrated in Figure~\ref{fig:conceptual}. We aim to exploit the advances in large pre-trained foundation models for image segmentation, The Segment Anything Model (SAM)~\cite{kirillov2023sam}, which offers strong generalizable semantic features. In this work, we use the SAM feature encoder to guide the construction of better cost volumes for MVS. We also propose a SAM-style decoder to extract semantic maps and use the depth predicted by the MVS branch as a dense decoder prompt. Our experimental results show improved performance from the use of these depth cues.

We perform an extensive quantitative and qualitative evaluation on the large ScanNet~\cite{dai2017scannet} dataset of indoor scenes. We demonstrate considerable improvement over monocular multi-task methods (11\% relative improvement in the semantic mIoU metric, and more notably, 40\% improvement in the absolute depth error). Our method also performs better than the competing single-task MVS methods (with a 10\% relative improvement in depth prediction over CasMVSNet), and better than the competing single-task segmentation approaches (17\% relative improvement over AdapNet++\cite{valada2020adapnetplusplus}), matching the performance of RGB-D algorithms, while requiring only RGB inputs.

Summarizing our contributions, we present:
\begin{itemize}
\itemsep-0.25em 
\item[1)] A unified architecture that simultaneously solves the Multi-View Stereo and the Semantic Segmentation problems, filling the gap between monocular and full 3D multi-task methods, a first to our knowledge;
\item[2)] A method to enrich MVS-based cost volumes for depth prediction using rich SAM segmentation features;
\item[3)] A semantic decoder that takes advantage of both the rich SAM features, and the predicted depth;
\item[4)] Quantitative and qualitative studies of the proposed approach on the ScanNetv2 dataset show consistent improvement over both multi-view depth prediction and segmentation models and over algorithms that jointly predict depth and segmentation from a single image.
\end{itemize}

\begin{figure*}
  \centering
  \includegraphics[width=\linewidth]{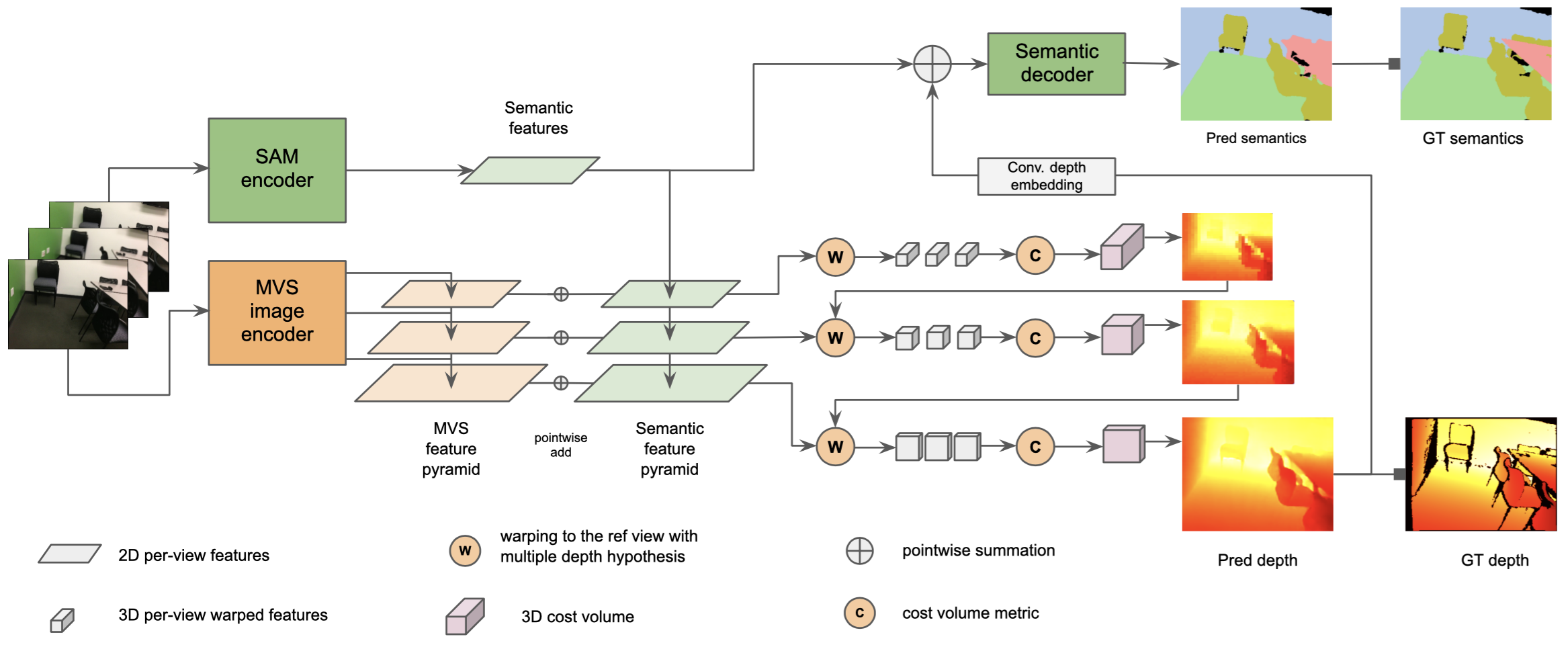}
  \caption{Architecture overview. Given a set of input RGB views (one reference and several other source images), we produce a depth map, and a semantic segmentation map for the reference view. The model first extracts per-view 2D features fusing the outputs of the convolutional MVS encoder and a transformer-based semantic encoder. The 2D features are warped onto the reference view with an array of depth hypotheses, and a 3D cost volume is constructed at a range of scales, where the initial depth prediction from a coarser scale allows to increase the resolution of the hypothesis step at a finer scale, focusing on refinement around the established coarse value. A 3D convolutional network decodes the cost volume into a depth prediction at each scale. The final depth prediction is used as a dense prompt for the transformer-based semantic decoder. The model is supervised with the ground truth depth and semantic labels during training.}
  \label{fig:main}
    \vspace{-1em}
\end{figure*}

\section{Background and Related work}

\textbf{Monocular depth \& segmentation}
Monocular semantic segmentation and depth estimation both perform dense predictions from a single image. Semantic segmentation requires a class label for each pixel \cite{deeplabv1, deeplabv2, deeplabv3, valada2020adapnetplusplus, pspnet, segnet, ronneberger2015unet}, while depth estimation predicts depth values that lift the pixels to 3D points \cite{li2022depthformer, liu2023va, saxena2023monocular, agarwal2023attention, piccinelli2023idisc}. 
There have been works exploring sequential (one task guides the other) and joint learning of both tasks (multi-task). Casser \etal \cite{casser2019unsupervised} use precomputed instance segmentation masks to handle objects in highly dynamic scenes. And a range of works exists that simultaneously learn to predict segmentation and depth \cite{wang2015towards,wang2020sdc,he2021sosd,eigen2015predicting}. We compare with four monocular methods: MT RefineNet~\cite{nekrasov2019real}, PAD Net~\cite{xu2018padnet}, MTI Net~\cite{vandenhende2020mti} and MTAN~\cite{liu2019mtan}. PAD Net leverages a set of intermediate tasks to guide the final segmentation and depth predictions. MTAN proposes task-specific attention modules. However, monocular approaches are naturally limited by the information encoded in a single image and struggle to recover complete geometry, so in this work we focus on reconstruction from multi-view images.

% \subsection{Full 3D segmentation and reconstruction: sequential and joint}

\textbf{Full 3D segmentation and reconstruction}
To utilize information from multiple images, methods based on Truncated Signed Distance Functions (TSDF) have been proposed which can perform real-time processing and fusion of a sequence of images or a video. Atlas~\cite{murez2020atlas} and NeuralRecon~\cite{sun2021neuralrecon} use back-projection to lift 2D features into 3D space and regress a TSDF in a voxel grid. Atlas~\cite{murez2020atlas} performs a running average aggregation and produces a scene-level voxel grid, while NeuralRecon~\cite{sun2021neuralrecon} predicts fragments of the scene and fuses them together through an RNN.

Segmentation and reconstruction can benefit from each other by taking additional depth information into account or by being guided by semantic information. Hane \etal \cite{hane2013joint} simultaneously solve the tasks of dense 3D scene reconstruction from multiple images and image semantic segmentation by utilizing class-specific smoothness assumptions in place of standard smoothness assumptions. 3DMV~\cite{dai20183dmv} utilizes the reconstruction of a RGB-D scan to predict a 3D semantic segmentation.

While these methods achieve high quality reconstruction and segmentation, they require many frames to work on and sometimes other inputs such as an initial mesh reconstruction. Our work aims at jointly learning depth and segmentation prediction from just a few views.

\textbf{Multi-View Stereo (MVS)}
% Talk about MVS methods.
The goal of Multi-View Stereo (MVS) is to reconstruct 3D scenes from multiple posed RGB images, usually by predicting depth maps and then fusing them together. 
% Modern approaches perform either prediction of the depth map for each frame, or learn explicit volumetric scene representation. 
One widely used approach builds plane-sweep volumes \cite{yao2018mvsnet, gu2020casmvsnet, ding2022transmvsnet, huang2018deepmvs, zhao2023mvpsnet} on top of depth hypotheses. Many SOTA methods are built upon MVSNet \cite{yao2018mvsnet}, which consists of extracting image features, warping features using homography to construct 3D cost volumes and applying a 3D CNN to estimate depth. CasMVSNet \cite{gu2020casmvsnet} builds the cost volumes in a cascaded way to improve time and memory efficiency. TransMVSNet \cite{ding2022transmvsnet} further proposes to improve the feature extraction and matching task by leveraging a transformer to aggregate long-range context information within and across images.

Semantic segmentation has been deployed to improve MVS quality in a range of works. 
 Semantic features can be used as a consistency constraint that favors the reconstructions yielding a consistent segmentation across binocular \cite{yang2018segstereo} or multi-view images \cite{xu2021selfsupervisedcoseg}. They also help with 3D plane fitting in textureless regions where photometric feature matching often fails \cite{yuan2023tsar, stathopoulou2021semanticallyderived}.
 
However, to the best of our knowledge, no existing method jointly learns MVS and segmentation in a multi-task manner. Our method simultaneously estimates depth and semantic maps. We show that the two tasks benefit from each other. Furthermore, only a few MVS methods \cite{yang2022mvs2d} report their performance on indoor datasets, while others report their performance on table-top datasets (see Figure~\ref{fig:dtu_scannet_views}). We benchmark our method on ScanNet \cite{dai2017scannet}.

\textbf{Foundation models for segmentation}

The Transformer model was initially proposed for natural language processing (NLP)~\cite{agarwal2023attention} introducing the attention mechanism to infer dependencies between language tokens. Transformers were then adopted for vision tasks achieving great success especially with the availability of large datasets~\cite{dosovitskiy2020vit,liu2021swin}. Vision Transformer (ViT) \cite{dosovitskiy2020image} is now a widely used model, which splits an image into patches and applies a Transformer on the encoding of a sequence of image patches. Although ViT was first introduced for image classification, it can serve as a rich feature extraction backbone on a range of vision tasks, similar to its convolutional counterparts like ResNet~\cite{he2016resnet}. 

Producing truly universal features for segmentation was not straightforward. In this work, we leverage the Segment Anything Model (SAM) \cite{kirillov2023sam} that was recently proposed as a foundation model for open-set segmentation. We use SAM to extract semantic features to construct cost volumes that are more efficient for depth prediction. SAM has ViT as its image encoder and shows good zero-shot performance on several tasks such as edge detection and instance segmentation inside a given box. SAM's mask decoder operates on the image embeddings and accepts prompts of two types: sparse (e.g. foreground and background points of the object of interest), and dense (e.g. coarse masks). We follow this style of the decoder and adapt it to our needs, so that MVS depth predictions serve as dense prompts and learned class embeddings as sparse prompts.

%%%%%%%%%%%%%%%%%%%%%%%%%%%%%%%%%%%%%%%%%%%%%%%%%%%%%%%%%%%%%%%%%%%%%%%%5
%%%%%%%%%%%%%%%%%%%%%%%%%%%%%%%%%%%%%%%%%%%%%%%%%%%%%%%%%%%%%%%%%%%%%%%%5
%%%%%%%%%%%%%%%%%%%%%%%%%%%%%%%%%%%%%%%%%%%%%%%%%%%%%%%%%%%%%%%%%%%%%%%%5
%%%%%%%%%%%%%%%%%%%%%%%%%%%%%%%%%%%%%%%%%%%%%%%%%%%%%%%%%%%%%%%%%%%%%%%%5
%% Method

\section{Method}
\label{sec:method}

We consider the problem of joint depth prediction and segmentation of a scene from a set of $M$ RGB images. Given input images ${I_i, \forall i=1,...,M}$ of spatial size $W \times H$ and corresponding camera poses $P_i = \{K_i, (R_i, t_i)\}$ (intrinsic parameters $K_i$, rotation matrix $R_i$ and translation vector $t_i$), we aim to predict a pixel-wise depth map $D_i$ and a segmentation map $S_i$. We follow the Multi-View Stereo (MVS) problem set-up, where we consider one of the $M$ images as a reference image and use the remaining images to predict the depth map and segmentation map for the reference image. We repeat this $M$ times by choosing each one of the $M$ images as the reference image. Note that this is different from monocular depth map and segmentation estimation techniques where each image is used in isolation for prediction.

We take inspiration from the state-of-the-art MVS architecture for depth prediction, CasMVSNet~\cite{gu2020casmvsnet}, and for segmentation, Segment Anything (SAM)~\cite{kirillov2023sam}. We propose a novel approach, as shown in Figure~\ref{fig:main}, that unifies these techniques to improve both depth and segmentation prediction. The model extracts 2D features from individual views through the geometry-aware convolutional MVS encoder, as well as the semantic-aware transformer-based SAM encoder, and fuses these features. Section~\ref{sec:method:2d_encoders} describes the 2D feature pipeline in detail. A multiscale cost volume is constructed from these per-view features and decoded into a depth probability volume, as discussed in Section~\ref{sec:method:mvs}. The depth prediction of the MVS branch is used as a dense prompt for the transformer-based semantic decoder (Section~\ref{sec:method:semantic_decoder}). The model is then trained end-to-end for joint depth prediction and semantic segmentation.

\subsection{2D encoders and feature fusion for MVS}
\label{sec:method:2d_encoders}

The first step of an MVS approach extracts deep geometry-aware features that are good for matching across $M$ input views. We design this MVS-specific 2D encoder, as shown in Figure~\ref{fig:main}, following CasMVSNet~\cite{gu2020casmvsnet}.  This encoder follows the design in~\cite{lin2017fpn} producing a feature pyramid with increasing spatial resolutions at 3 scales. 

We augment the geometry-aware features of the MVS encoder with semantics-aware features. The Segment Anything Model (SAM)~\cite{kirillov2023sam} was very recently proposed as a foundation model for open set segmentation. It uses the Vision Transformer (ViT)~\cite{dosovitskiy2020vit} as a feature encoder.  Given a high resolution (normally, with a long side of $1024$ pixels)  RGB image at the input, the ViT encoder breaks it into tokens that have a spatial size of $16 \times 16$ pixels, producing a grid of tokens. The transformer is applying multiple layers of non-local attention, keeping the spatial resolution fixed throughout all layers, producing a single rich segmentation-aware feature tensor. We note that the spatial resolution of these features roughly corresponds to the coarsest resolution of the MVS pyramid, and we resize the tensor with bilinear interpolation to that resolution. We additionally create a lightweight convolutional pyramid encoder for the SAM features to match the spatial and feature dimensions at all three scales of the MVS pyramid.

The geometry-aware MVS features and the SAM semantic features are fused together with a simple pointwise summation at all scales of the pyramid, same as it is done in other works (e.g. in FuseNet~\cite{hazirbas2017fusenet} for geometry-aware depth features and the visual RGB features). In our preliminary experiments other approaches, such as stacking didn't show any performance improvement. 

\subsection{Cost volumes and depth prediction}
\label{sec:method:mvs}

Once the features are extracted from individual views, a range of depth hypotheses are created for each pixel of the reference view $i=0$. Features from source view $i$ are warped with each depth hypothesis $d$ using a homogeneous coordinate mapping: 

\begin{equation}
    \begin{pmatrix}
    u' \\ v' \\ 1
    \end{pmatrix}
    \sim 
    K_i R_i^T \left(
    R_0 K_0^{-1} d
    \begin{pmatrix}
    u \\ v \\ 1
    \end{pmatrix}
    + t_0 - t_i
    \right)
\end{equation}

This mapping creates per-view 3D feature volumes $V_i$ in the reference coordinate view. All $V_i$ are aggregated into a cost volume $C$ using the variance-based cost metric:

\begin{equation}
    C = \dfrac{1}{M} \sum\limits_{i=1}^M \left( V_i - \overline{V} \right)^2,
\end{equation}

where $\overline{V}$ is the average feature volume. Such a 3D cost volume is created at a range of scales, as proposed in CasMVSNet~\cite{gu2020casmvsnet}, and decoded into the single-channel 3D depth probability volume $P$ with 3D convolutions. When the probability volume is normalized, it defines a distribution over the range of depth hypothesis, and the depth prediction for the reference frame at pixel $(i,j)$ is inferred as:

\begin{equation}
    D_0^{i,j} = \sum\limits_{t} P^{i,j,t} d_t\,.
\end{equation}

At the coarsest scale, the entire depth range is covered for some fixed $[d_{min}, d_{max}]$ with equally spaced hypotheses $d_t$. Once the coarser depth is predicted at one scale, it is interpolated to a finer scale to serve as a mean of the new (narrower) depth hypothesis range, allowing for higher resolution of the hypothesis interval. The final depth prediction $D_0$ is performed at the finest scale that has the same spatial $W \times H$ resolution as the input image.

\subsection{Semantic segmentation}
\label{sec:method:semantic_decoder}

We use the SAM-pretrained encoder in Section~\ref{sec:method:2d_encoders} and modify the mask decoder since our semantic segmentation task is different from the open-set instance segmentation in SAM. The mask decoder transforms the input image embedding and a set of prompt embeddings into a set of output masks. As a prediction is required for each of the $K$ semantic classes, we introduce $K$ learned embeddings to serve as prompts, analogous to SAM's \emph{sparse prompts}.

Additionally, the depth maps are available from the MVS prediction branch, which we use similarly to SAM's \emph{dense prompts}. A dense embedding is created with a shallow convolutional network that aligns with the image feature representation and is summed point-wise with the semantic encoder's features as shown in Figure~\ref{fig:main}. We note that other learned dense prompts could be used as well in addition to our depth maps. For example, one may think of using the SAM's zero-shot edge maps, however such an approach would be very computationally expensive (SAM needs to decode 768 masks and run NMS and Sobel filter post-processing to produce the edge map). We limit this study to only our predicted depth prompt.

\begin{figure}[ht]
  \centering
  \includegraphics[width=\linewidth]{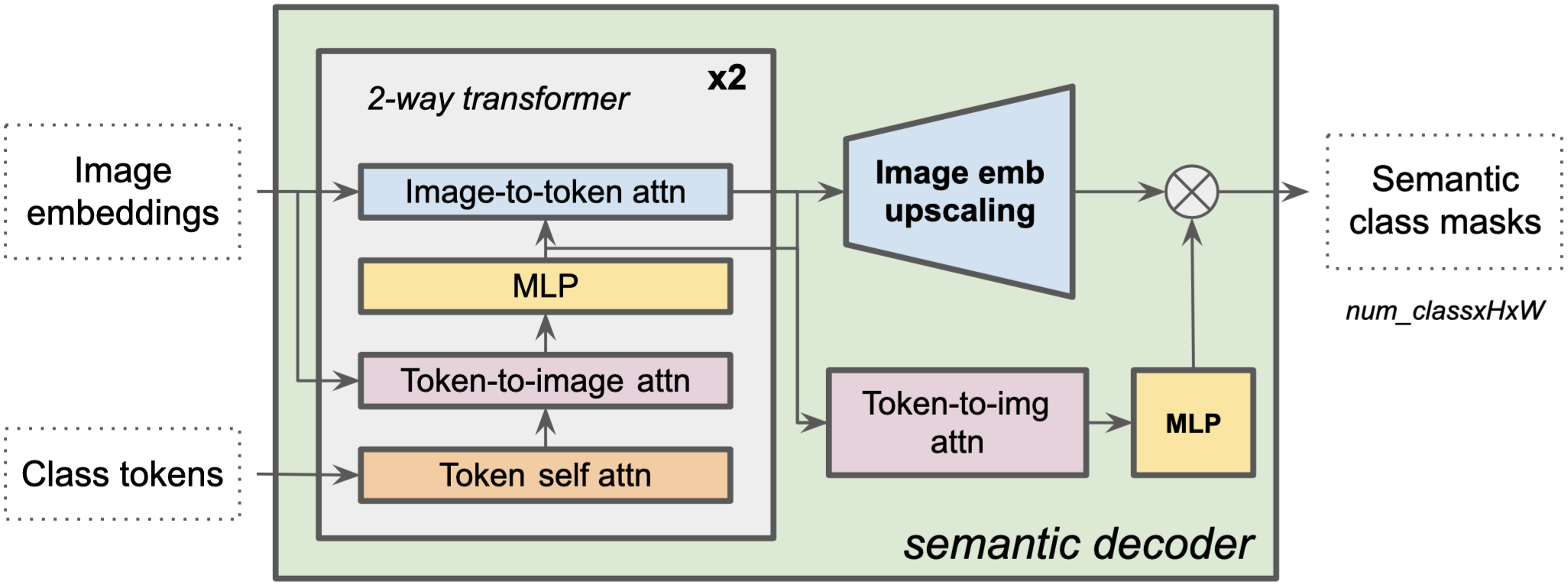}
  \caption{Semantic decoder. Two-way attention blocks are used to simultaneously update the class tokens and the image embeddings. Image embeddings are then upscaled, while the final MLP transforms the class tokens into queries that decode the semantic segmentation masks. Decoding is done by correlating the queries with the final image embeddings at each spatial location.}
  \label{fig:semantic_decoder}
\end{figure}

The decoder is represented by a lightweight two-way transformer that communicates information between the grid of image tokens and a set of queries, followed by a dynamic mask prediction block. It is depicted in Figure~\ref{fig:semantic_decoder}.  Each of the two transformer blocks starts with token self-attention, followed by token-to-image cross-attention (tokens serving as attention queries), a point-wise MLP that updates the tokens, and finally image-to-token cross-attention (with image embeddings serving as attention queries). Thus, each of the transformer blocks updates both the tokens (entangling tokens via self-attention and aggregating image features through token-to-image attention), and the image features (infusing the tokens information through image-to-token attention). Standard residual connections and layer normalizations are employed after each of the four steps in each block that are omitted from Figure~\ref{fig:semantic_decoder} to avoid clutter. Image features are also supplied with positional encodings that are added pointwise at the input of each attention layer.

The resulting image features are upscaled with a shallow 2-layer deconvolutional network to match the $W \times H$ resolution of the system's input image, and tokens are additionally embedded with a token-to-image attention block and an MLP to produce the semantic class queries. Semantic masks are decoded through a cross-product between the image embeddings and the class queries.

% Hey misha, if you haven't submit a backup, you should get down to 8 pages and submit one now

\textbf{Loss function} We minimize the joint loss function:
\begin{equation}
    L = L_{seg} + \alpha L_{MVS},
    \label{eq:total_loss}
\end{equation}
where $L_{seg}$ is a cross-entropy semantic segmentation loss, and $L_{MVS}$ is a smooth L1 depth loss.

\section{Experiments}

\begin{table}
  \centering
  \begin{tabular}{@{}lccc@{}}
    \toprule
    Method & \#views & Abs (cm)~$\downarrow$ & Rel (\%)~$\downarrow$ \\
    \midrule
    MVS2D~\cite{yang2022mvs2d} & 3 & 10.8 & 5.9 \\
    TransMVSNet~\cite{ding2022transmvsnet} & 3 & 13.2 & 8.3\\
    CasMVSNet*~\cite{gu2020casmvsnet} & 3 & 10.0 & 6.0 \\
    Ours, MVS + UNet feat & 3 & 9.5 & 5.8 \\
    Ours, MVS + SAM feat & 3 & 9.3 & 5.5 \\
    Ours, full  & 3 & 9.0 & 5.6 \\
    \midrule
    Ours, full & 5 & 8.3 & 5.2 \\
    \bottomrule
  \end{tabular}
  \caption{Multi-view depth estimation. We follow the same cost volume-based cascaded architecture as CasMVSNet. "MVS + UNet features" and "MVS + SAM features" are single-task depth prediction models that augment the 2D MVS geometric features with semantic features (see Sec~\ref{sec:method:2d_encoders}) and do not have a semantic decoder. Our full model jointly predicts depth and semantic segmentation and fine-tunes the last layers of the SAM encoder.}
  \label{tab:depth_estimation}
\end{table}

\subsection{Training settings}

Models are trained with the AdamW optimizer ($\beta_1 = 0.9$, $\beta_2 = 0.999$, $weight\_decay=10^{-2}$), a learning rate of $10^{-3}$, and a total batch size of 8 distributed over 4 GPUs, for 16 epochs. With most of the SAM encoder parameters frozen, we optionally tune the last $0-3$ blocks of the SAM encoder with a $10\times$ smaller learning rate (i.e. $10^{-4}$). The input image size is set to $320\times256$, and both depth and segmentation maps are predicted at this resolution. We use a smooth L1 loss between the ground truth and the predicted depth with $\beta=0.02$ (i.e. the threshold is set at the error value of 2~cm), ignoring values outside of the depth range. Semantic and depth loss contribute equally ($\alpha=1$ in Eq.~\ref{eq:total_loss}).

\subsection{Architecture settings}

ViT-base is used as a semantic feature extractor for SAM.

A cascade of cost volumes is constructed to predict depth at 3 scales. The depth is predicted across the range of $[0.1m, 5.0m]$, assuming $192$ virtual depth bins. This corresponds to the base interval of roughly $2.5cm$ between the depth hypotheses. The network spaces the intervals with a ratio of $[4\times, 1\times, 0.5\times]$ of the base interval at its three scales, evaluating $[48, 32, 8]$ depth hypotheses at each pixel in the corresponding scales. Unless explicitly specified, our networks use $M=3$ input RGB views.

\subsection{Datasets}

We evaluate our results on the ScanNet~\cite{dai2017scannet} dataset. ScanNet is a large-scale established indoor recognition benchmark. It contains 1513 scans across 707 locations. Semantic segmentation tasks includes 20 categories. For most of our studies (e.g. Tables~\ref{tab:depth_estimation},~\ref{tab:multitask_monocular},~\ref{tab:semantic_seg_ablations}) we use the train/validation split provided by MVS2D~\cite{yang2022mvs2d} to benchmark the results of the methods on the same sets of views. For other experiments, the official split of the ScanNet v2 scenes into training and validation sets is used.

\subsection{Metrics}

We use the standard metrics for evaluating the predicted depth maps~\cite{eigen2014depth}: mean absolute depth error (Abs err), mean relative depth error (Rel err), root mean square error (RMSE). Absolute depth error and RMSE can are in centimeters, relative depth error is measured with respect to the ground truth. Lower is better for all depth evaluation metrics. Segmentation outputs are evaluated by the mean intersection-over-union (mIoU). Higher is better.

\subsection{Depth prediction}

\begin{table}
  \centering
  \begin{tabular}{@{}lcccc@{}}
    \toprule
    Method & mIoU~$\uparrow$  & Abs~$\downarrow$  & Rel~$\downarrow$  & RMSE~$\downarrow$  \\
    \midrule
    MT RefineNet~\cite{nekrasov2019real} & 42.0 & 32.4 & 21.7 & 44.3 \\
    PAD Net~\cite{xu2018padnet} & 42.2 & 26.9 & 17.4 & 37.8 \\
    MTI Net~\cite{vandenhende2020mti} & 53.7 & 22.4 & 14.0 & 32.2 \\
    MTAN~\cite{liu2019mtan} & 56.1 & 22.7 & 14.6 & 32.6 \\
    \midrule
    Ours, 3 views & 62.5 & 9.0 & 5.6 & 14.9 \\
    \bottomrule
  \end{tabular}
  \caption{Comparisons to monocular multitask methods for joint depth prediction and semantic segmentation.}
  \label{tab:multitask_monocular}
\end{table}

CasMVSNet~\cite{gu2020casmvsnet} and TransMVSNet~\cite{ding2022transmvsnet} serve as our MVS baseline models for depth prediction. We use the train/validation split released in MVS2D~\cite{yang2022mvs2d} that includes both the scene partitions, and the view sampling, in order to perform a fair comparison of the $M$-view MVS methods against each other. MVS2D is trained using the parameters reported in~\cite{yang2022mvs2d}. The paper's codebase is publicly available and results are reported on ScanNet. CasMVSNet and TranMVSNet are trained using our custom settings described above. In Table~\ref{tab:depth_estimation} we analyze the benefits of using semantic features to augment the geometric MVS features during cost volume construction. Our baseline CasMVSNet has an absolute depth error of 10.0, and MVS2D shows as 10.8. TransMVS, although reported as a superior model on the standard MVS datasets~\cite{jensen2014dtu,knapitsch2017tanks}, shows a worse generalization to ScanNet with an absolute depth error of 13.2.

We first demonstrate the benefit of using standard encoder-decoder semantic features. A U-Net~\cite{ronneberger2015unet} with a ResNet~\cite{he2016resnet} backbone is pre-trained on the same ScanNet train split, and the three scales of the decoder pyramid that correspond to the MVS decoder spatial scales are used instead of our SAM encoder in Figure~\ref{fig:main}. The semantic prediction heads are discarded in this experiment, and all the remaining U-Net weights are tuned during the training, improving the depth prediction performance to 9.5.

Augmenting 2D features with the SAM encoder output brings the error down to 9.3cm even without any encoder tuning, while training our full model with the segmentation branch and tuning of the last 3 SAM encoder layers demonstrates an additional improvement with an error of 9.0. Note that although the segmentation decoder outputs are not influencing the MVS branch during inference, during training the gradients still propagate back to the SAM encoder features, serving as an additional regularizer. Note that the the evaluation is performed against the ground truth depth maps that are also noisy and imprecise, and thus the lower bound for a method's performance is greater than zero, and unknown, as discussed below in Section~\ref{ex:qualitative}. Interpreting the errors' absolute values is therefore not straightforward without looking at some qualitative results.

%%%%%%%%%%%%%%%%%%%%%%%%%%%%%%%%%%%%

In Table~\ref{tab:multitask_monocular} we compare with monocular multitask methods for joint depth prediction and semantic segmentation with open-source implementation. We re-train the methods on our train/val split of ScanNet to report the results. We notice that the results are similar to those reported in the papers for NYUD2, since the datasets have similar data distributions. For example, Multi-Task RefineNet~\cite{nekrasov2019real} reports 42.02~mIoU and 56.5~RMSE on NYUD2, while our performance on ScanNet shows 42.0~mIoU and 44.3~RMSE.

\begin{table}
  \centering
  \begin{tabular}{@{}lccc@{}}
    \toprule
    Method & Modality & \#views & mIoU~$\uparrow$  \\
    \midrule
    SegNet \cite{segnet} & RGB & 1 & 27.5 \\
    DeepLab v2 \cite{deeplabv2} & RGB & 1 & 43.9 \\
    DeepLab v3 \cite{deeplabv3} & RGB & 1 &  50.1 \\
    AdapNet++~\cite{valada2020adapnetplusplus} & RGB & 1 & 53.0 \\
    \midrule
    FuseNet~\cite{hazirbas2017fusenet} & RGB-D & 1 & 63.8 \\
    EMSAFormer \cite{fischedick2023efficient} & RGB-D & 1 & 63.8 \\
    SSMA~\cite{valada2020adapnetplusplus} & RGB-D & 1 & 67.4 \\
    \midrule
    Ours & RGB & 5 & 62.1 \\
    \bottomrule
  \end{tabular}
  \caption{2D semantic segmentation on ScanNet validation set.}
  \label{tab:semantic_seg}
  \vspace{-1em}
\end{table}

\begin{table}
  \centering
  \begin{tabular}{@{}lcc@{}}
    \toprule
    Method & Modality & mIoU~$\uparrow$ \\
    % Aka SAM-simple in the code
    UNet (ResNet-34) & RGB & 50.9 \\
    UNet (ResNet-50) & RGB & 34.5 \\
    SAM (frozen) + conv decoder & RGB & 27.2 \\
    SAM (frozen) + our decoder & RGB & 47.2 \\
    SAM (tuned) + our decoder & RGB & 59.0 \\
    Our full model & RGB & 62.5  \\
    \midrule
    SAM (frozen) + our dec. (GT depth) & RGB-D & 52.4 \\ 
    SAM (tuned) + our dec. (GT depth) & RGB-D & 63.4 \\
    \bottomrule
  \end{tabular}
  \caption{2D semantic segmentation ablations. Conventional methods such as UNet overfit easily when only using ImageNet-pretrained backbones. SAM pretrained features are much more robust, but produce poor performance when decoded using convolutional layers. Using a transformer-based decoder improves the quality considerably, especially with encoder feature tuning. Our full model uses the depth predicted from the MVS branch and comes close to matching the performance of the RGBD version that uses ground truth depth maps as a decoder prompt.}
  \label{tab:semantic_seg_ablations}
\end{table}

\begin{table}
  \centering
  \begin{tabular}{@{}lccc@{}}
    \toprule
    Method & Abs (cm)~$\downarrow$ & Rel (\%)~$\downarrow$ & mIoU~$\uparrow$ \\
    \midrule
    CasMVSNet & 10.0 & 6.0 & NA \\
    % corresponds to Ours, SAM features in Table 1
    MVS + SAM features & 9.3 & 5.5 & NA \\
    % Corresponds to SAM (frozen) + our decoder
    Semantic SAM (frozen) & NA & NA & 47.2 \\
    Semantic SAM (tuned) & NA & NA & 59.0 \\
    Ours, no SAM$\rightarrow$MVS & 10.0 & 6.1 & 60.0 \\
    Ours, full & 9.0 & 5.6 & 62.5 \\
    
    \bottomrule
  \end{tabular}
  \caption{Model ablations. CasMVSNet is the baseline depth prediction approach. MVS + SAM features is our model without the segmentation head (i.e. \emph{semantic features help depth prediction}). Semantic SAM is our semantic decoder on top of the SAM encoder with no depth prediction and no depth prompting. "Ours, no SAM$\rightarrow$MVS" cuts the SAM to MVS feature connection, i.e. depth prediction is performed without SAM features, while still running segmentation with the depth prompt.}
  \label{tab:depth_ablations}
    \vspace{-1em}
\end{table}

\newcolumntype{C}{>{\centering\arraybackslash}m{6.0em}}
\begin{table*}
  \centering
  % \begin{tabular}{@{}lccc@{}}
  \begin{tabular}{l*6{C}@{}}
    \toprule
    & scene0076\_00
    % & scene0170_01
    & scene0390\_00
    & scene0449\_00
    & scene0465\_00(1)
    & scene0465\_00(2)
    & scene0465\_00(3)
    \\
    \midrule

    Target view
    & \includegraphics[width=6em]{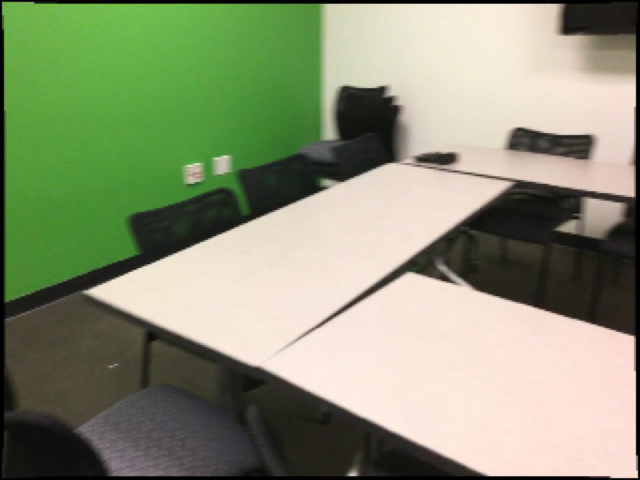}
    % & \includegraphics[width=6em]{scene0170_01_frames_1120_1100_1140.png.view0.png}
    & \includegraphics[width=6em]{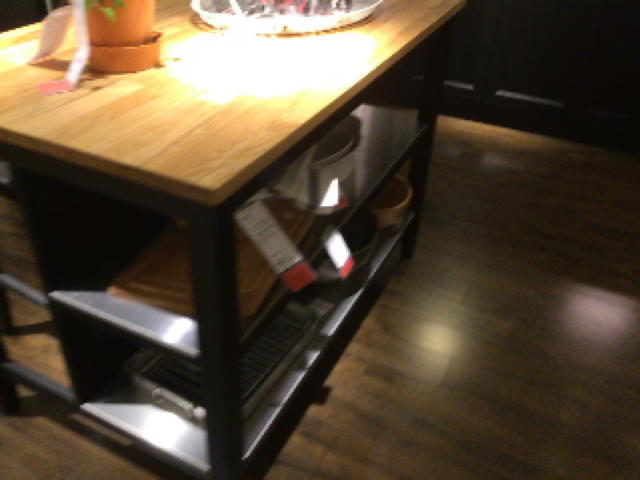}
    & \includegraphics[width=6em]{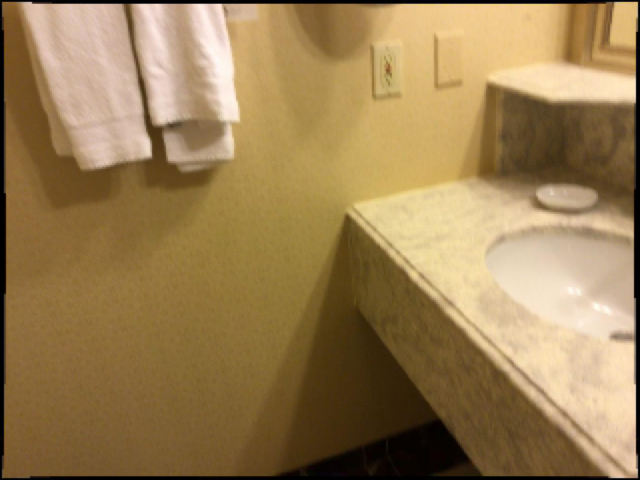}
    & \includegraphics[width=6em]{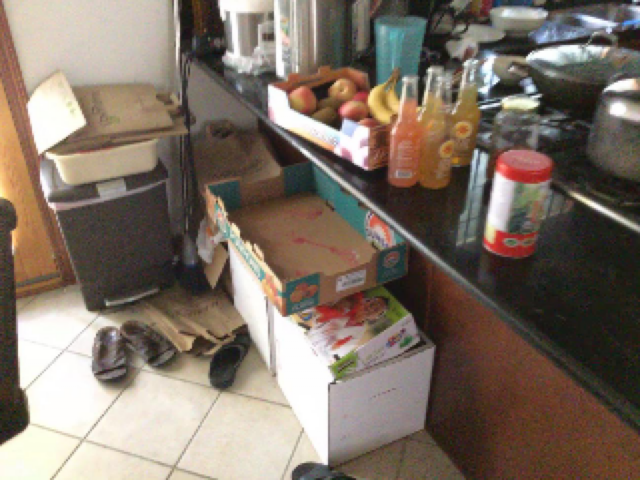}
    & \includegraphics[width=6em]{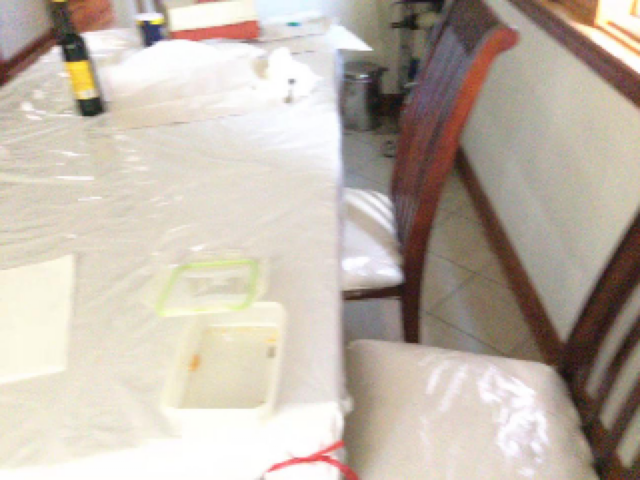}
    & \includegraphics[width=6em]{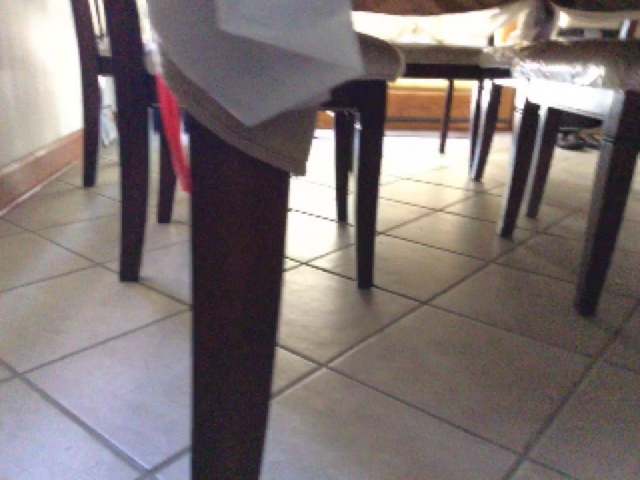}
    \\

    \midrule

    MTAN~\cite{liu2019mtan}
    & \includegraphics[width=6em]{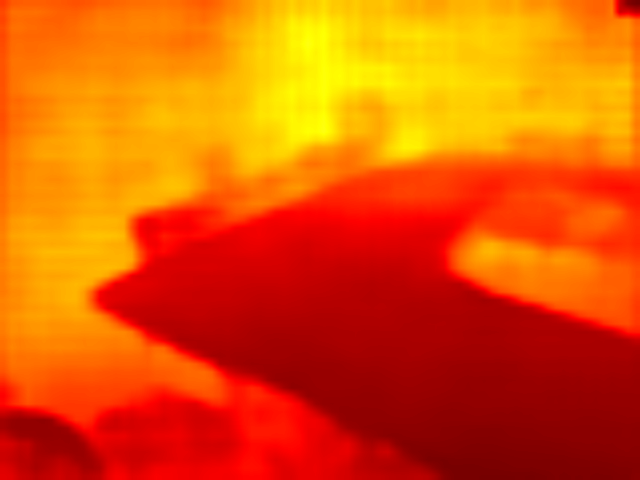}
    & \includegraphics[width=6em]{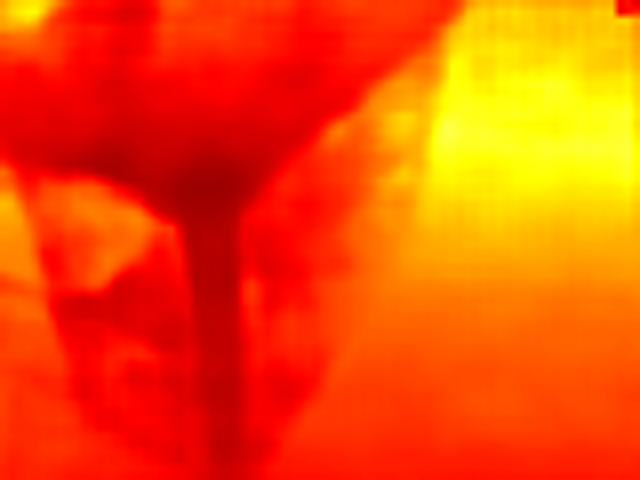}
    & \includegraphics[width=6em]{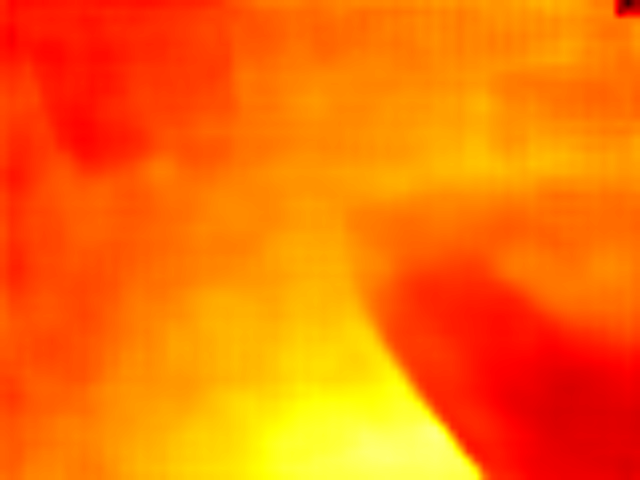}
    & \includegraphics[width=6em]{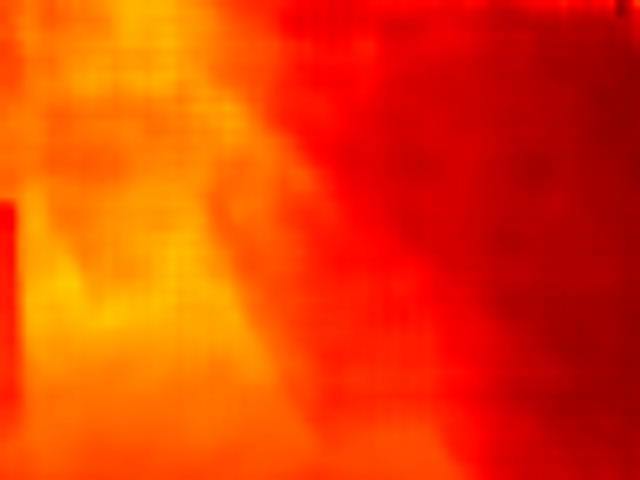}
    & \includegraphics[width=6em]{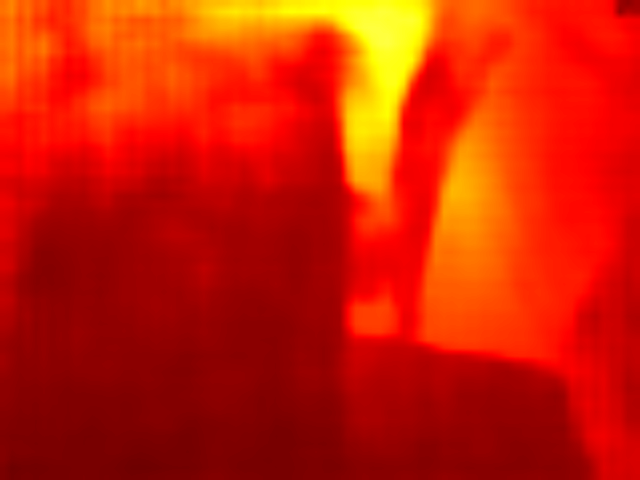}
    & \includegraphics[width=6em]{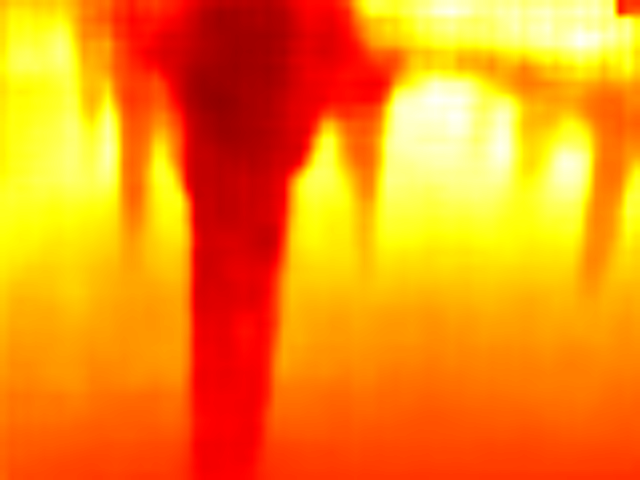}
    \\
    
    CasMVSNet~\cite{gu2020casmvsnet}
    & \includegraphics[width=6em]{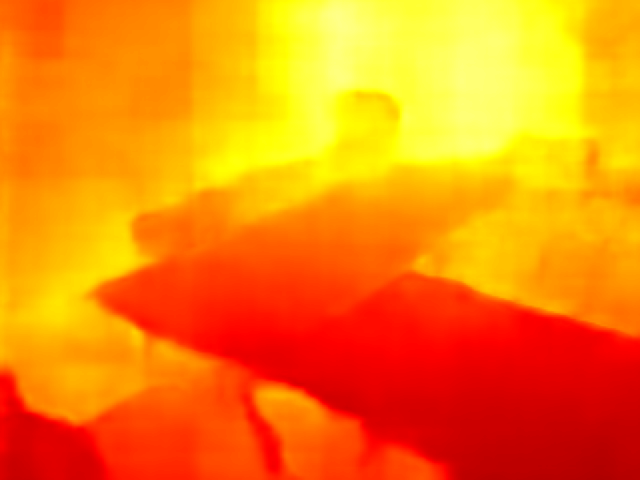}
    %& \includegraphics[width=6em]{cas_scene0170_01_frames_1120_1100_1140.png.depth_pred.png}
    & \includegraphics[width=6em]{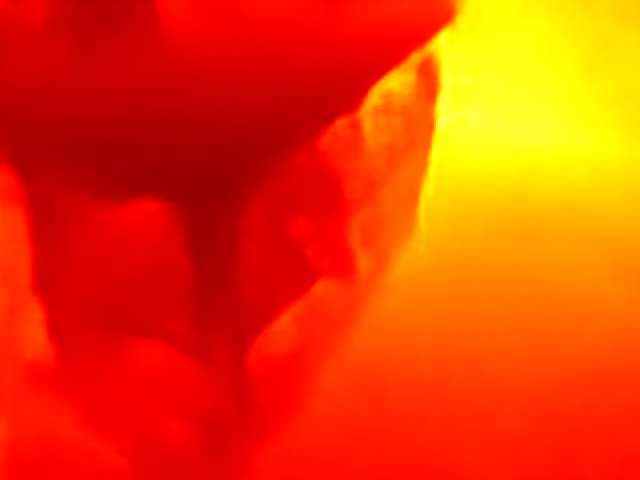}
    & \includegraphics[width=6em]{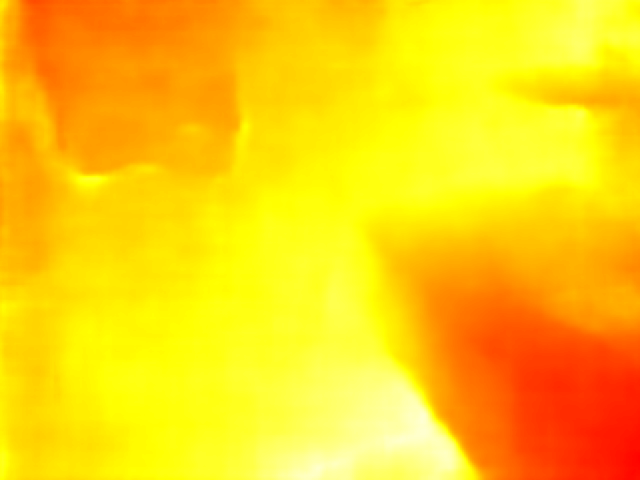}
    & \includegraphics[width=6em]{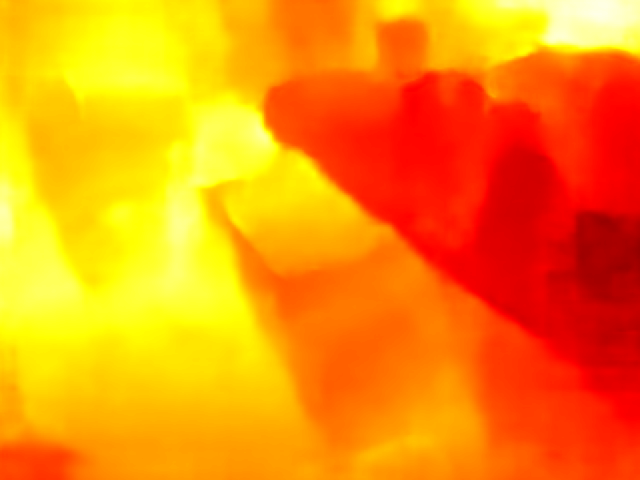}
    & \includegraphics[width=6em]{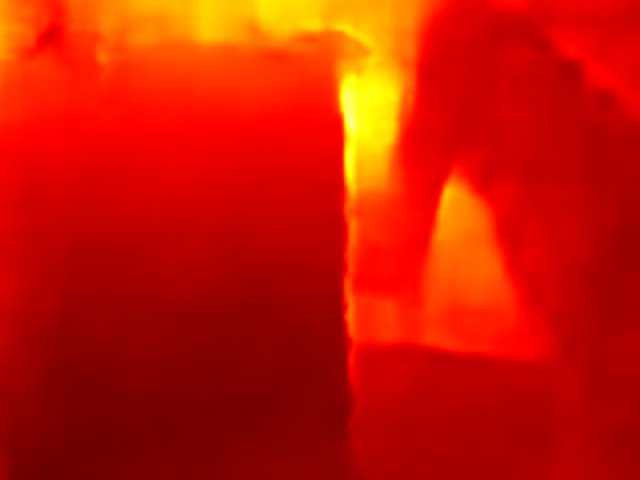}
    & \includegraphics[width=6em]{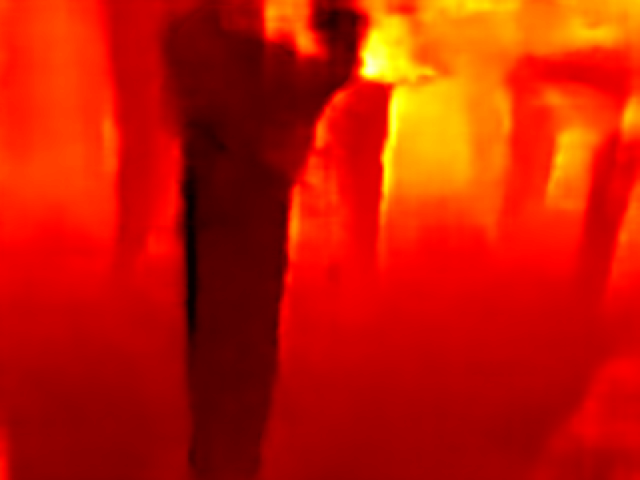}
    \\

    Ours
    & \includegraphics[width=6em]{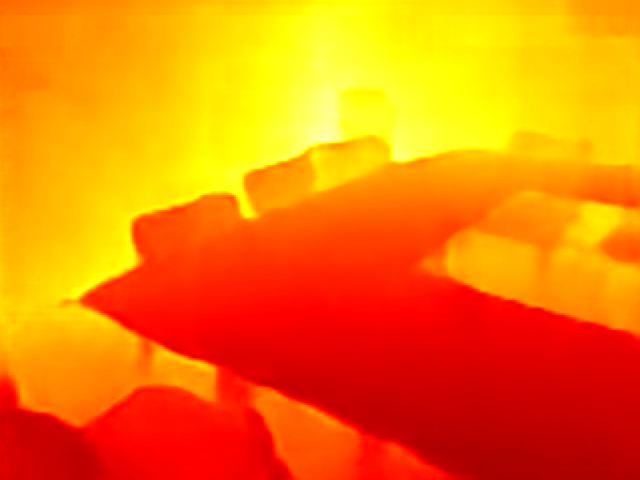} 
    %& \includegraphics[width=6em]{cas_sam_scene0170_01_frames_1120_1100_1140.png.depth_pred.png}
    & \includegraphics[width=6em]{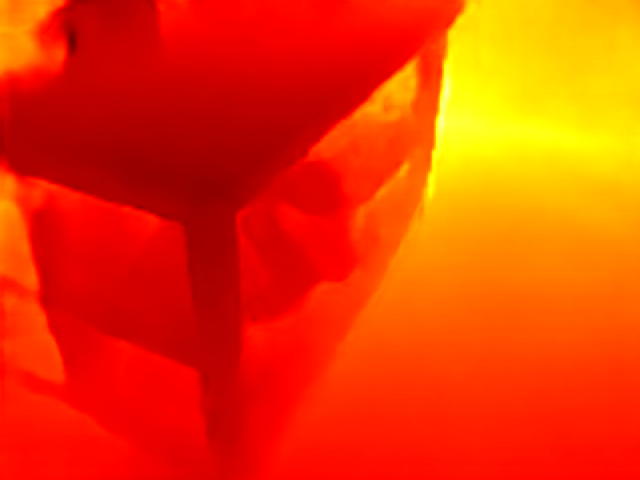}
    & \includegraphics[width=6em]{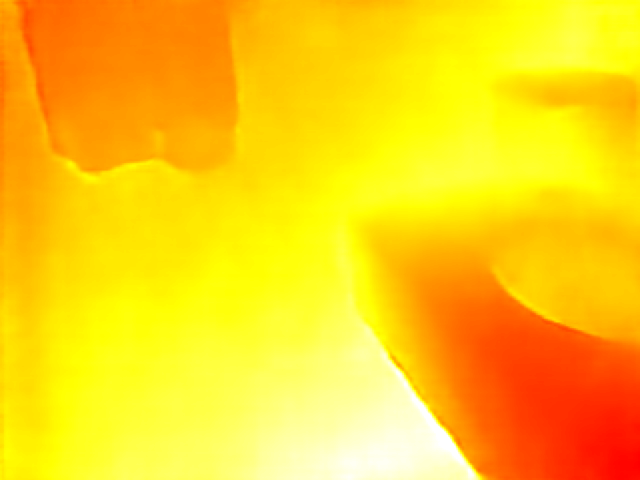} 
    & \includegraphics[width=6em]{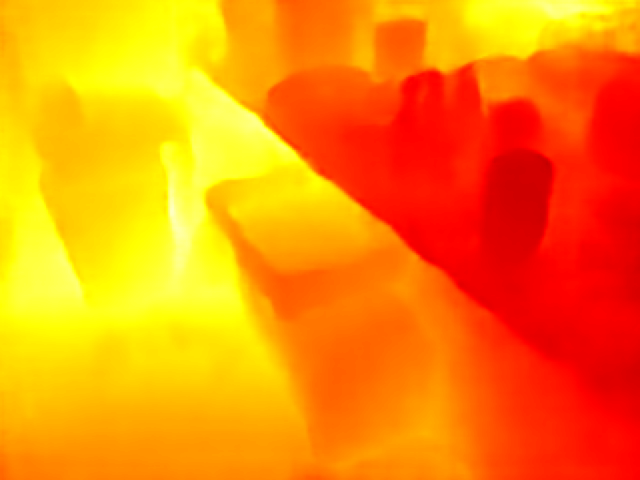} 
    & \includegraphics[width=6em]{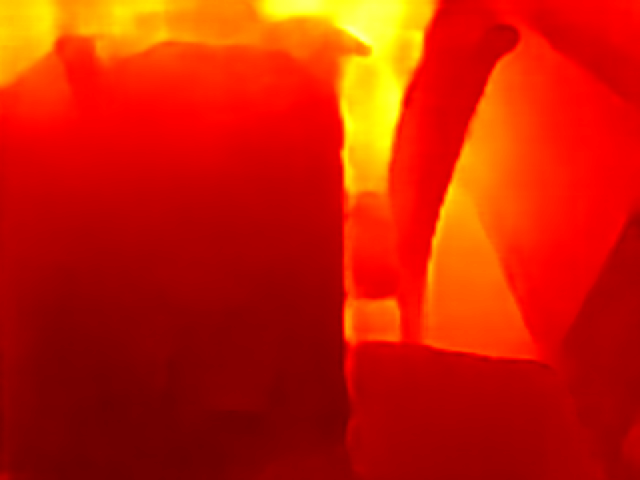}
    & \includegraphics[width=6em]{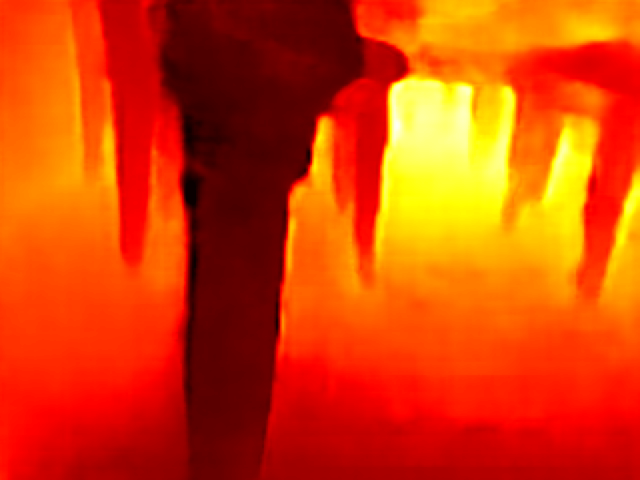}
    \\

    GT depth
    & \includegraphics[width=6em]{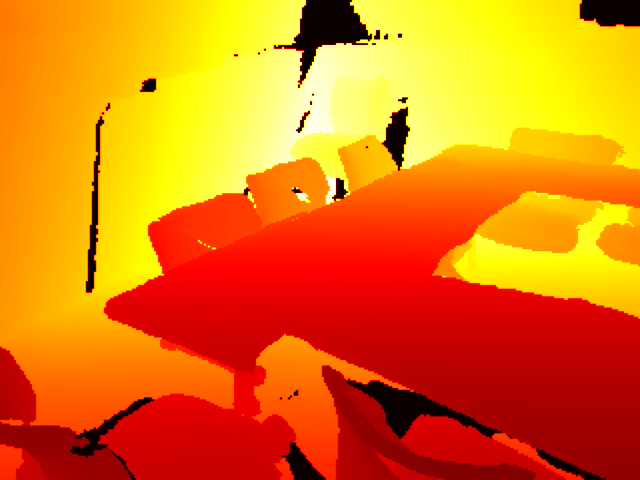} 
    %& \includegraphics[width=6em]{scene0170_01_frames_1120_1100_1140.png.depth_gt.png}
    & \includegraphics[width=6em]{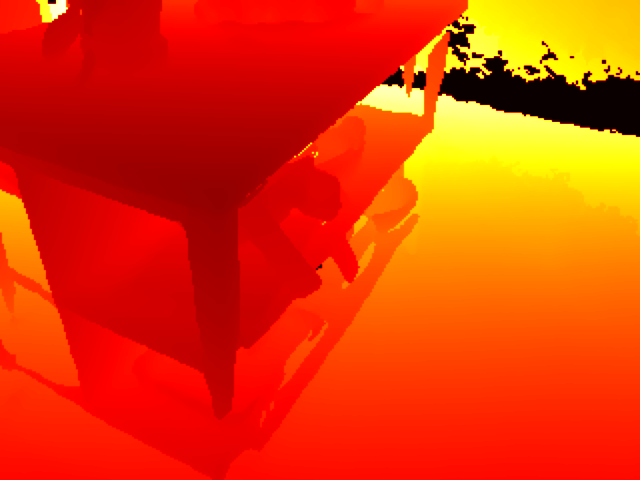}
    & \includegraphics[width=6em]{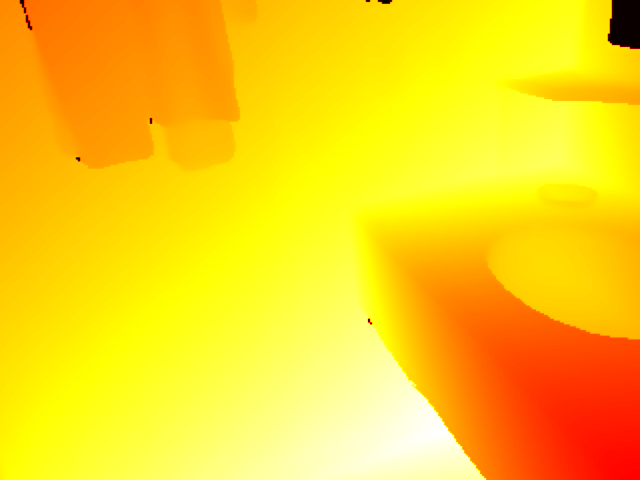}
    & \includegraphics[width=6em]{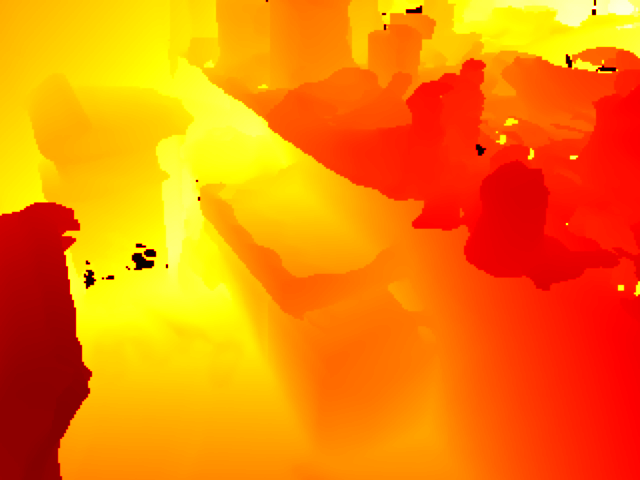}
    & \includegraphics[width=6em]{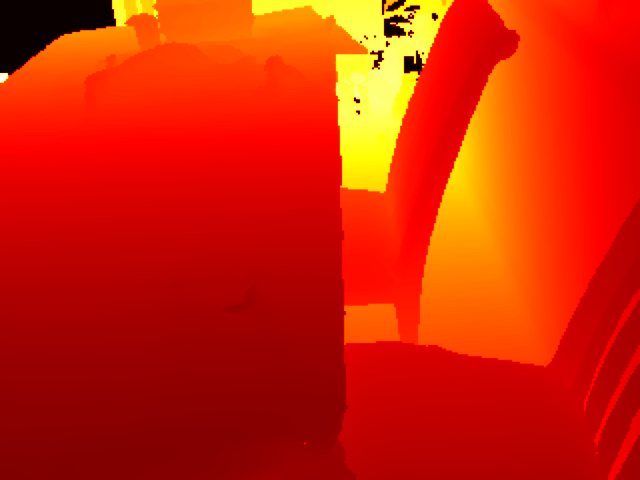}
    & \includegraphics[width=6em]{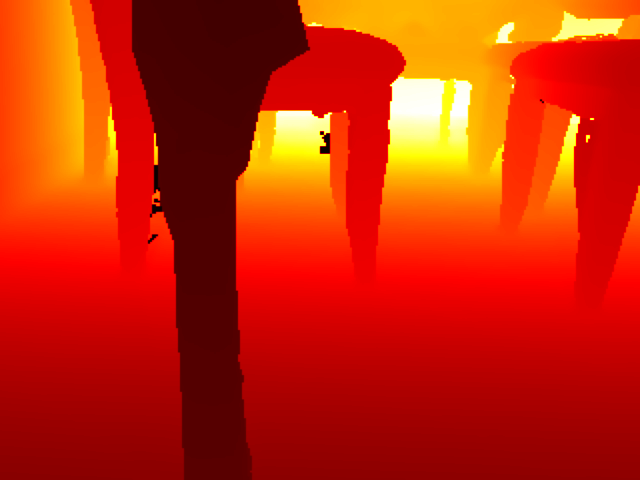}
    \\

    \bottomrule
  \end{tabular}
  \caption{Depth estimation results on ScanNet. Existing MVS techniques, MTAN and CasMVSNet, perform poorly on textureless flat surfaces like walls (scene0076\_00, scene0449\_00) and tables (scene0390\_00,scene0465\_00(2)) and on thin structures with strong occlusions like legs of tables (scene0390\_00,cene0465\_00(2)) and clutter on kitchen counter top (scene0465\_00(1)). In contrast, our proposed technique produces significantly better performance on these regions by leveraging semantic understanding of the scene. Both CasMVSNet and Ours are trained using $M=3$ views on the MVS2D~\cite{yang2022mvs2d} split.} % (i.e. the listed scenes are not part of the training set, even though they come from the full scannetv2\_train benchmark set).}
  \label{tab:depth_qualitative}
  \vspace{-1em}
\end{table*}

\subsection{Qualitative studies}
\label{ex:qualitative}

Table~\ref{tab:depth_qualitative} shows the depth estimation results for three methods: multitask monocular MTAN~\cite{liu2019mtan}, multi-view CasMVSNet~\cite{dai2017scannet}, and our model. Both CasMVSNet and our model are trained and tested using only $M=3$ views. While from Table~\ref{tab:multitask_monocular} the improvement of our method is obvious compared to monocular methods, there is still a question of how much we really improve over the baseline MVS in terms of depth prediction (i.e. is 9.0 Ours vs 10.0 CasMVSNet in Table~\ref{tab:depth_estimation} a considerable difference or not). As we already stated above, due to the fact that the ground truth is imprecise, the lower bound for such metrics as the absolute depth error are nonzero and unknown. Comparing our method to the baseline MVS visually we notice that it performs consistently better across the dataset, with a considerable difference in some cases. In the first column the effect on flat matte structures, such as walls, is pronounced, where semantic cues may help to recover the known MVS failures. 
In the last column, an unconventional view and the glossy floor surface make it hard to infer the depth from geometric patterns. Semantic meaning identifies the objects as table and chair legs, making it easier to determine boundaries.

\subsection{Segmentation}

Table~\ref{tab:semantic_seg_ablations} offers a study of our SAM-based semantic segmentation. We first establish a benchmark with UNet-type encoder-decoder architectures using ResNet as a backbone. We notice that deeper models overfit on ScanNet since the data variety is limited. In particular, the ResNet-34 based method outperforms the deeper ResNet-50 version. The SAM-pretrained transformer proves to be much more robust. However, a simple convolutional decoder delivers poor performance (27.2 mIoU). The transformer-based decoder described in Section~\ref{sec:method:semantic_decoder} significantly improves the mIoU to 47.2 even with the frozen encoder, while additional feature tuning reaches the quality of 59.0 mIoU. In our full model that jointly predicts depth and segmentation maps, the predicted depth is used as a dense prompt to the decoder, for which the performance of 62.5 mIoU is reported in the table. The methods marked as RGB-D in Table~\ref{tab:semantic_seg_ablations} use the ground truth depth as the input. Our multi-view RGB model gets close to matching the corresponding RGB-D setting.

\subsection{Ablation studies}

Table~\ref{tab:depth_ablations} summarizes the effects of individual components of our framework. CasMVSNet sets the absolute depth error baseline of 10.0 without the use of any semantic information; our Semantic SAM is trained in isolation (with no MVS branch and no depth prompt) setting the segmentation baseline at 47.2 mIoU and 59.0 with encoder tuning. MVS + SAM features represent our model without the semantic head, improving the depth prediction. Our full model shows an improvement in both depth prediction (9.0 absolute error) and semantic segmentation (62.5 mIoU).

\vspace{-0.5em}
\section{Conclusion}
\vspace{-0.5em}

In this work, we introduced a novel approach to joint depth and segmentation estimation in the multi-view setting, the first to our knowledge. We show that jointly estimating depth and segmentation is better than separately predicting them in multiview settings in Table 1 and 3 respectively. This discovery in itself is not new and has been previously explored for multi-task learning in single-view monocular depth and segmentation estimation. However, we show that multi-task learning in multiview is significantly better than single-view approaches in Table 2, highlighting the importance of this work to extend multi-task learning from single-view to multi-view. Multi-view-based 3D scene understanding approaches that require only 3-5 images are significantly more practical in many robotics applications compared to approaches that require a full-dense 3D scan of the room.

%%%%%%%%% REFERENCES
\newpage
{\small
\bibliographystyle{ieee_fullname}
\bibliography{egbib}
}

\end{document}